\documentclass[preprint,12pt,3p]{elsarticle}

\makeatletter
\def\ps@pprintTitle{%
 \let\@oddhead\@empty
 \let\@evenhead\@empty
 \def\@oddfoot{}%
 \let\@evenfoot\@oddfoot}
\makeatother

\usepackage{graphics}
\usepackage{graphicx}
\usepackage{amssymb}
\usepackage[linesnumbered,lined,boxed,commentsnumbered]{algorithm2e}
\usepackage{afterpage}
\usepackage{float}
\usepackage{hyperref}

\begin{document}

\begin{frontmatter}

\title{Deep Neural Network for \\Real-Time Autonomous Indoor Navigation}

\author{Dong Ki Kim}
\author{Tsuhan Chen}
\address{Cornell University}

\begin{abstract}
Autonomous indoor navigation of Micro Aerial Vehicles (MAVs) possesses many challenges. One main reason is because GPS has limited precision in indoor environments. The additional fact that MAVs are not able to carry heavy weight or power consuming sensors, such as range finders, makes indoor autonomous navigation a challenging task. In this paper, we propose a practical system in which a quadcopter autonomously navigates indoors and finds a specific target, i.e. a book bag, by using a single camera. A deep learning model, Convolutional Neural Network (ConvNet), is used to learn a controller strategy that mimics an expert pilot's choice of action. We show our system's performance through real-time experiments in diverse indoor locations. To understand more about our trained network, we use several visualization techniques. 
\end{abstract}

\end{frontmatter}

\section{Introduction}
\label{sec1}
\begin{figure}[ht]
  \includegraphics[scale=1.5]{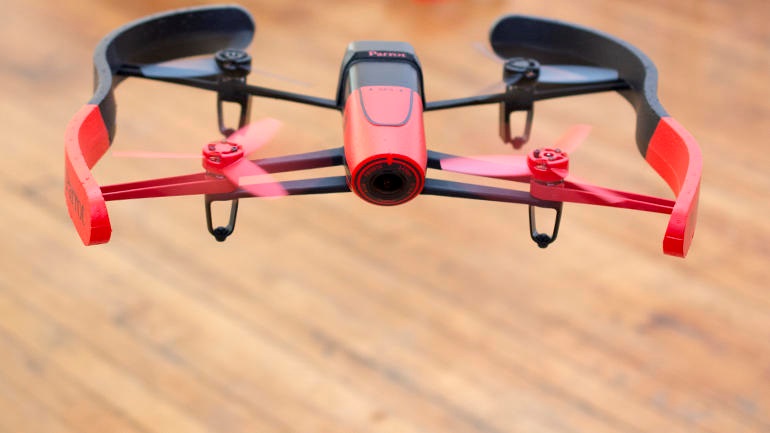}
  \centering
  \caption{A MAV used for our research, Parrot Bebop Drone \cite{bebop}}
  \label{fig:bebop}
\end{figure}

Micro Aerial Vehicles (MAVs), such as quadcopters equipped with a camera (Figure \ref{fig:bebop}), are widely used in many applications, such as rescue, exploration, and entertainment. In recent years, outdoor autonomous navigation has been successfully accomplished through the use of global positioning system (GPS) \cite{gps}. However, GPS shows limited precision in indoor environments, which brings many challenges for indoor autonomous flight.\par

Several solutions have been proposed for indoor autonomous navigation. One solution is a Simultaneous Localization and Mapping (SLAM). Using laser range finders, RGB-D sensors, or a single camera, a 3-D map of unknown indoor environments and its position in the map can be inferred for autonomous flight (\cite{slam1}-\cite{slam3}). Another solution is based on stereo vision. By computing disparity between stereo images, depth can be estimated (\cite{stereo1},\cite{stereo2}). SLAM, however, is not practical for MAVs because building a 3D model is computationally heavy. Additionally, the constructed 3D structure often does not perform well in environments with devoid of trackable features (e.g., walls). Depth estimated by stereo vision shows low performance in texture-less regions and can suffer from specular reflections. The additional fact that most of publicly available quadcopters have only one built-in camera makes the solutions not practical. \par

In this paper, we present a practical system enabling a quadcopter to navigate autonomously indoors and find a specific target, i.e., a book bag. Our system does not require any range finding sensors, but only a single camera. Our approach is to train a classifier that mimics an expert pilot’s choice of action. A deep learning model, Convolutional Neural Network (ConvNet), is used to train the classifier with our custom dataset. Our classifier consistently receives a purely visual input from a quadcopter and returns a flight command that best mimics the expert’s action. Through our real-time test experiments, we show that our system correctly finds a target while concurrently performing autonomous navigation with a success rate of 70-80\%.\par

One advantage of our approach is that it prevents MAVs from colliding into a wall. This advantage could supplement SLAM: when it fails to localize its position due to the devoid of trackable environments, our approach can be used instead, until it is certain about its position. Our system also does not construct a 3-D map, so it has relatively less computation. Additionally, our approach does not require high resolution cameras, which makes our system attractive for many widely distributed MAVs.\par

In the discussion section, we use visualization techniques (\cite{yosinski},\cite{deconvnet}) to visualize features and representations learned by our classifier. Through the visualization, we examine what features our classifier learned from training and which feature affected the classification performance.\par

This paper makes three principal contributions. First, we introduce a practical system based on deep learning for autonomous indoor flight. The system is practical because it requires only a single camera and is computationally less expensive. Second, we provide our custom dataset. The dataset is composed of 7 indoor locations, and each location has its own unique appearance. The diversity of our dataset would be useful for other indoor research. Third, we visualize the trained deep model by visualization techniques. The visualization adds knowledge for understanding the model. The remainder of this paper describes these contributions in detail. Section 2 introduces related works about autonomous navigation and flight. Section 3 explains our hardware platform, dataset, and training details. Section 4 demonstrates our test experiments. Section 5 includes discussion about our deep learning model using visualization techniques. Finally, section 6 offers concluding remarks.

\newpage
\section{Related Works}
There has been impressive research on autonomous navigation and flight of MAVs. In this section, we examine related works.\par

\textbf{Range Sensor}: One possible solution is to use range sensors, such as laser range finders, infrared sensors, or RGB-Depth sensors. Bry et al. \cite{bry} presented a state estimation method using an on-board laser range finder and inertial measurement unit and showed aggressive flight in GPS-denied environments. Roberts et al. \cite{roberts} used one ultrasonic sensor and four infrared sensors and showed fully autonomous flight, i.e., collision avoidance. The range sensor, however, is not practical to most of publicly available quadcopters as the on-board device is often too heavy for MAVs and consume lots of power. Our work is based on only a monocular camera, which consumes a low power and is built in to most of quadcopters.\par

\textbf{SLAM}: Using range sensors or visual sensors, a 3-D map of unknown indoor environments can be inferred, while simultaneously estimating its position in the map (\cite{slam1},\cite{slam2},\cite{slam3}). Bachrach et al. \cite{bachrach} used a laser rangefinder sensor for a high-level SLAM implementation and exploring unknown indoor environments. Celik et al \cite{celik} presented autonomous indoor navigation based on SLAM using monocular vision. However, SLAM is computationally expensive due to the 3-D reconstruction. This causes unacceptable delay between perception and action. Also, SLAM shows low accuracy when it is applied to indoor environments, like walls, which contain insufficient feature points that can be tracked frame to frame. Our system does not perform path-planning. Thus our approach is closely related to minimizing the delay by reacting fast to its currently faced situation. Our system also shows robust performance on detecting and avoiding walls.\par

\textbf{Stereo Vision}: Accurate depth estimation and relative position estimation are possible using stereo cameras (\cite{stereo1},\cite{stereo2}). However, stereo vision algorithms suffer when they are used in texture-less regions, as it is hard to match features in one image to the corresponding features in the other image. The additional fact that most of publicly available quadcopters have only one built-in camera makes the solution not practical to a public. Our system shows robust performance in texture-less environments.\par

\textbf{Other Approach}: Other approach uses vanishing points. Bills et al. \cite{bills} used a monocular camera and found vanishing points. The points were used to fly in corridor environments. For staircase environments, they found center of the staircase. A front-facing short-range sensor, however, was additionally used to avoid collisions in corners and unknown environments. Our approach does not require the additional range sensor and can successfully perform collision avoidance with a monocular camera. Another approaches that are most closely related to our approach are approaches that learn control policies from input data. The ALVINN project \cite{alvinn} showed how the 3-layer artificial neural networks imitated a human driver’s response on road and performed autonomous vehicle driving. Ross et al. \cite{ross} applied a novel imitation learning strategy, the DAgger Algorithm, and learned a controller policy that imitated human pilot’s choice of action from demonstrations of the desired behavior. The system demonstrated a fast autonomous flight in natural forest environments. We extend these learning approaches and employ an advanced classifier, ConvNets, which learns to autonomously fly and finds a target based on purely visual input.

\section{Overview}
\begin{figure}[ht!]
  \includegraphics[width=\textwidth]{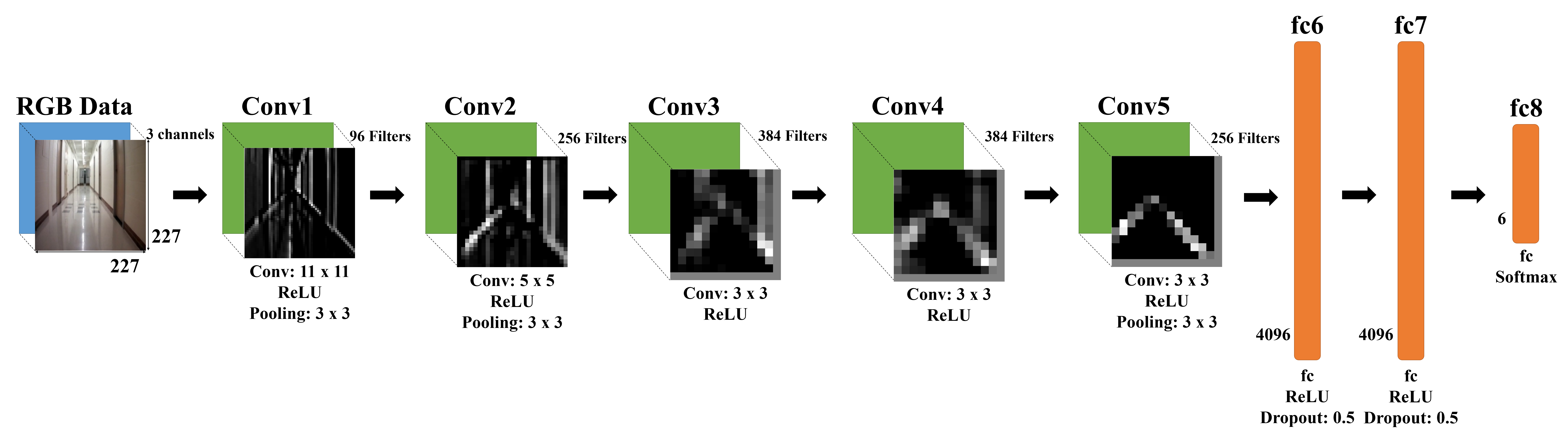}
  \centering
  \caption{Our classifier architecture. The architecture is similar to the CaffeNet architecture \cite{caffenet}, except that our last fully connected layer fc8 has six nodes. The input to the classifier is a RGB image, and the classifier outputs one of the six flight commands. The figure is best viewed in color and zoom-in electronic form.}
  \label{fig:network}
\end{figure}

The goal of our work is to learn a controller strategy that mimics an expert pilot’s choice of action. Given real-time images taken from MAVs, our trained classifier returns flight commands that best mimics the expert’s actions until a target is found, as outlined in Algorithm 1. The classifier is trained through an expert pilot demonstrating a desired flight given real-time images. Our training strategy allows the classifier to learn the most effective controller strategy with minimizing possible mistakes. A deep learning model of ConvNet is used as a classifier. We train the model by supervised learning. The architecture of our network has 5 convolutional layers and 3 fully connected layers (Figure \ref{fig:network}). The model’s parameters are learned through fine-tuning from the pre-trained CaffeNet model \cite{caffenet} with our custom dataset.

\vspace{1cm}

\IncMargin{1em}
\begin{algorithm}[H]
\SetKwData{Left}{left}\SetKwData{This}{this}\SetKwData{Up}{up}
\SetKwFunction{Union}{Union}\SetKwFunction{FindCompress}{FindCompress}
\SetKwInOut{Input}{Input}
\SetKwInOut{Prior}{Prior}
\SetKwInOut{Output}{Output}

\Input{Real-time image}
\Prior{Trained classifier}
\Output{Flight command}
    Initialize and take-off a drone\;
    \If{Take-off is complete}
    {
        Receive a new frame from a drone\;
        Input the new frame to the trained classifier\;
        Predict a command flight\;
        \eIf{Prediction confidence is low}
        {
            Hover a drone\;
            Go back to \small{3}\;
        }
        {
            \eIf{Predicted class is a target}
            {
                Stop (Land) a drone\;
                Break\;
            }
            {
                Return a corresponding flight command: Move Forward, Move Right, Move Left, Spin Right, or Spin Left\;
                Go back to \small{3}
            }
        }
    }
\caption{Autonomous Flight Process}
\end{algorithm}\DecMargin{1em}

\newpage
\section{Deep Neural Network for Autonomous Navigation}
In this section, we will first explain our hardware platform. Then we will demonstrate our custom dataset used for training ConvNet and details about training in the following section.

\subsection{Hardware Platform}
Our primary quadcopter is the Parrot Bebop Drone (Figure \ref{fig:bebop}). This quadcopter is currently available to the general public. The Bebop drone contains a single forward-facing camera, an ultrasound sensor for measuring ground altitude, and an on-board computer. Commands and images are exchanged via a WiFi connection between our host machine and the Bebop drone. The WiFi connection has a signal range of 250 meters (0.155 miles). From the Bebop’s stream connection, we receive an image of resolution of 640x368 pixels. We run our classifier on the host machine—a 2015 Alienware 15 (NVIDIA GeForce GTX 970M, Intel Core i5, 16GB memory), running Ubuntu 14.04.

\subsection{Dataset}
There exist many publicly available indoor images datasets (\cite{indoorDataset1},\cite{indoorDataset2}). Most of them, however, are not applicable to our approach because none of them do not provide ground-truth in flight commands. In addition, manually labeling flight commands by inferring could lead to a small mistake in a training dataset, which could potentially cause compounding errors. Thus creating our own dataset is necessary to achieve our goal.\par

\begin{figure}[h!]
  \includegraphics[scale=0.25]{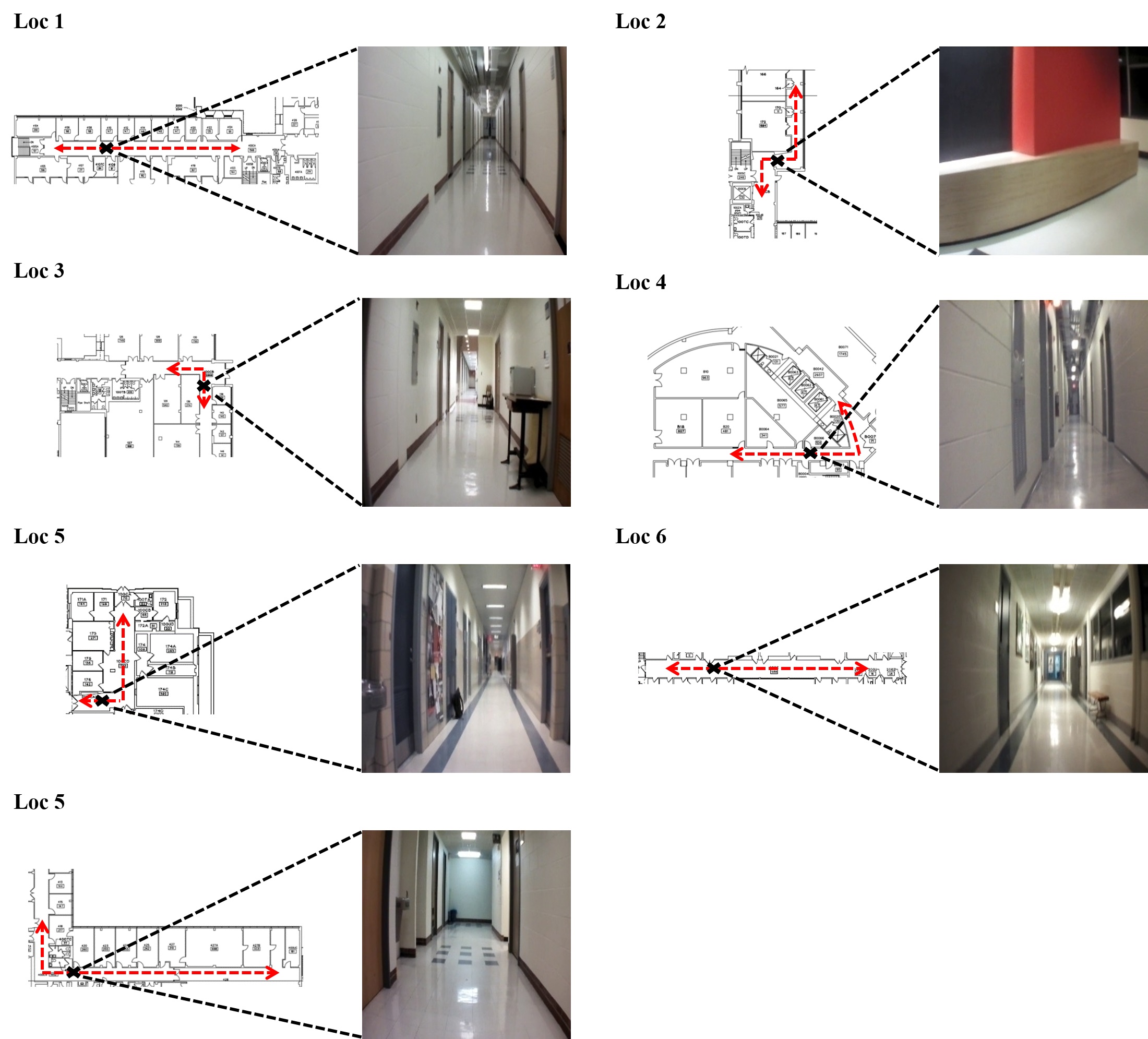}
  \centering
  \caption{Locations included in our training dataset. For each location, a sample image and its corresponding floor-plan are described.}
  \label{fig:trainingFloorPlan}
\end{figure}

Our dataset is composed of images collected from seven different indoor locations. The locations are either a corridor or corner. The selected environments have their own unique appearance, such as different building structure, brightness, and objects (i.e. desks). The locations are described in Figure \ref{fig:trainingFloorPlan}.\par

\begin{figure}[h!]
  \includegraphics[width=\textwidth]{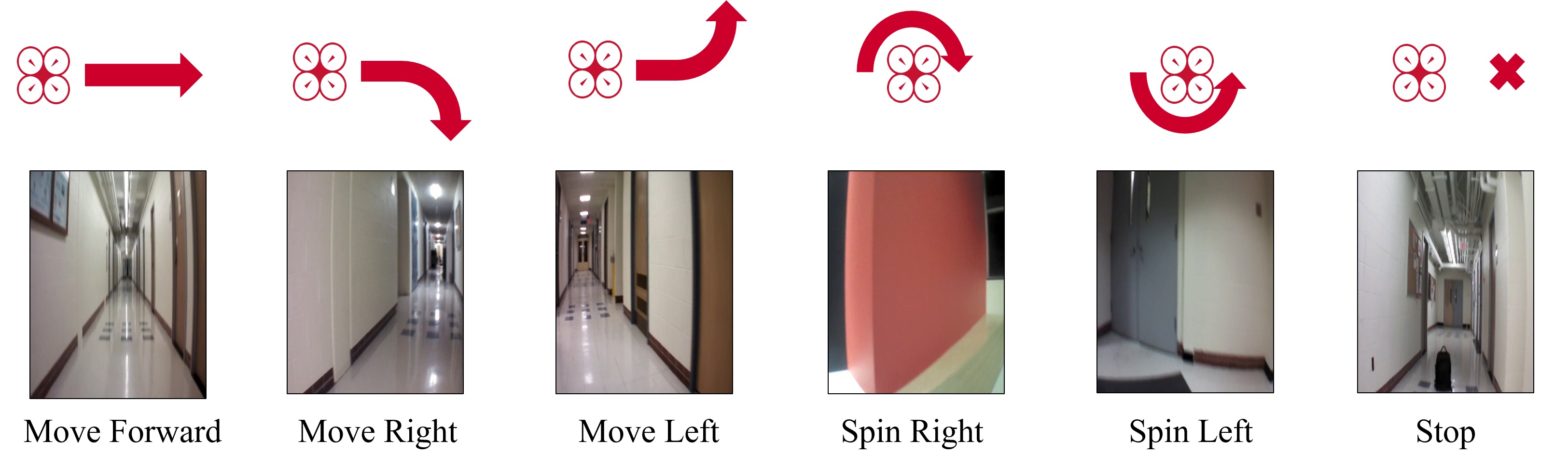}
  \centering
  \caption{Flight commands and corresponding sample images are shown.}
  \label{fig:flightCommand}
\end{figure}

\begin{figure}[h!]
  \includegraphics[scale=0.1]{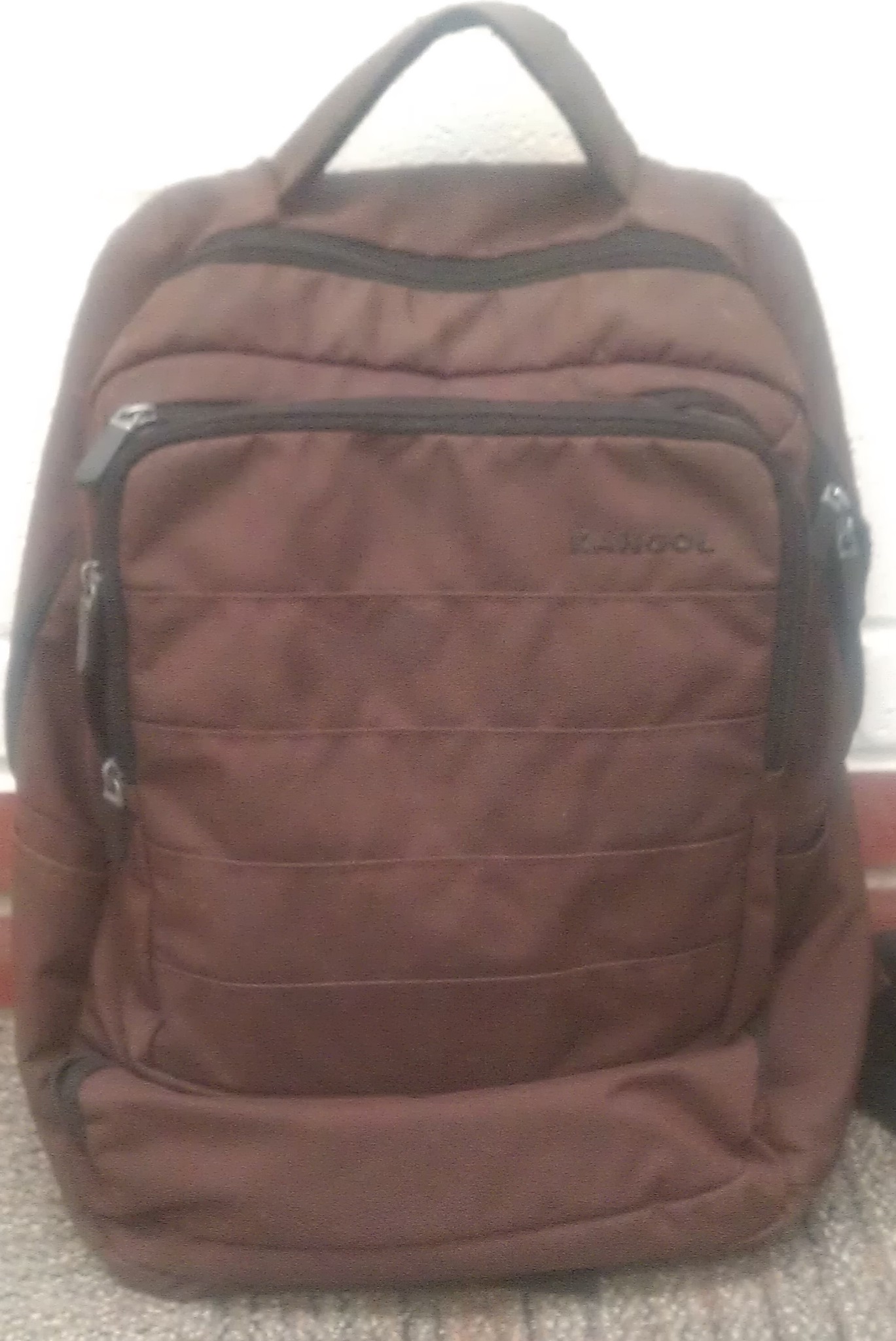}
  \centering
  \caption{A book bag used as the target.}
  \label{fig:bag}
\end{figure}

We notice that a constant height (i.e. 1 meter) is sufficient for flying in corridors or corners. Except for stairways, altering altitude is generally not important, and obstacles can be easily avoided using other directions (i.e. turning right or left). Furthermore, considering a constant height effectively reduces control complexity. We therefore control the quadcopter at a constant height of 1 meter using only six flight commands: Move Forward, Move Right, Move Left, Spin Right, Spin Left, and Stop (Figure \ref{fig:flightCommand}). The Stop flight command is for when the quadcopter finds a target. We set a target as a book bag (Figure \ref{fig:bag}).\par

\newpage
For each training location, an expert pilot controls the quadcopter using the six flight commands and demonstrates sample flights multiple times. We collect images taken from MAVs and corresponding flight commands for each demonstration. During the demonstrations, we cover as many as possible failure cases. For instance, when the quadcopter is close to colliding into walls, it should avoid the collision by turning to the opposite direction of its current propagation direction.\par

The images streamed from the Bebop drone have a noise. To train a classifier robust to noise, we generate additional images and augment our dataset size by adding Gaussian white noise of mean 0 and variance 0.01 to our dataset. The final dataset, therefore, has two times larger size than the previous dataset without noise.\par

Total images after data collection is summarized in Table \ref{table:totalDataset}. Please note that the dataset includes a large number of Move Forward flight command as moving forward is the most frequently used command, and it is a natural behavior to navigate corridors or corners. The dataset can be downloaded from: http://www.dongkikim.com/research/nav/index.html.

\begin{table}
\centering
\begin{tabular}{ |c || c c c c c c c || c| }
 \hline
 Flight Command & Loc 1 & Loc 2 & Loc 3 & Loc 4 & Loc 5 & Loc 6 & Loc 7 & Sum\\
 \hline
 Move Forward & 20000 & 6930 & 15000 & 7596 & 7330 & 20000 & 15000 & 91856\\
 Move Right & 2830 & 4000 & 2846 & 2076 & 1062 & 3094 & 2066 & 17974 \\
 Move Left & 2602 & 3790 & 3144 & 2828 & 614 & 2516 & 2100 & 17594\\
 Spin Right & 0 & 3382 & 306 & 834 & 468 & 0 & 360 & 5350\\
 Spin Left & 0 & 3468 & 340 & 648 & 460 & 0 & 222 & 5138\\
 Stop & 5210 & 4488 & 5162 & 5502 & 5596 & 4798 & 5876 & 36632\\
 \hline
\end{tabular}
\caption[Table caption text]{Number of training images taken from each location (Loc). Depending on the location type, either a corridor or corner, number of images for each flight command differs. For corridor environments, Move Forward images are collected more than the other flight commands. For corner environments, flight commands related to turning (Move Right \& Left, Spin Right \& Left) are collected more than the others.}
\label{table:totalDataset}
\end{table}

\subsection{Fine-tuning}
We start with the pre-trained CaffeNet model \cite{caffenet}. The CaffeNet model is composed of 5 convolutional layers and 3 fully connected layers. The model has similar architecture and performance to the AlexNet model with small differences, such as the order of pooling and normalization layers is switched. Because our system predicts one of the six flight commands, we replace the last fully connected layer fc8 with a layer composed of 6 nodes, as shown in Figure \ref{fig:network}. For the fine-tuning, we decrease the overall learning rate during training, but increase the learning rate of the newly introduced layer to allow the new layer learn faster than the rest of the model with our new data.\par

Please note that we disable mirror during training and testing. Similarly to the AlexNet model, the CaffeNet extracts 224 x 224 pixel sub-images from an 256 x 256 pixel image by mirroring. However, the mirroring does not apply to our approach: an image with Move Right flight command becomes an image that a quadcopter should move to left by mirroring, but the flipped image still has a label as Move Right.\par
We train over 20k iterations with a batch size of 255. With NVIDIA GTX 970M GPU and NVIDIA cuDNN \cite{cudnn}, the overall training time took 6 hours. 

\section{Experiments}
\begin{figure}[h!]
  \includegraphics[width=0.9\textwidth]{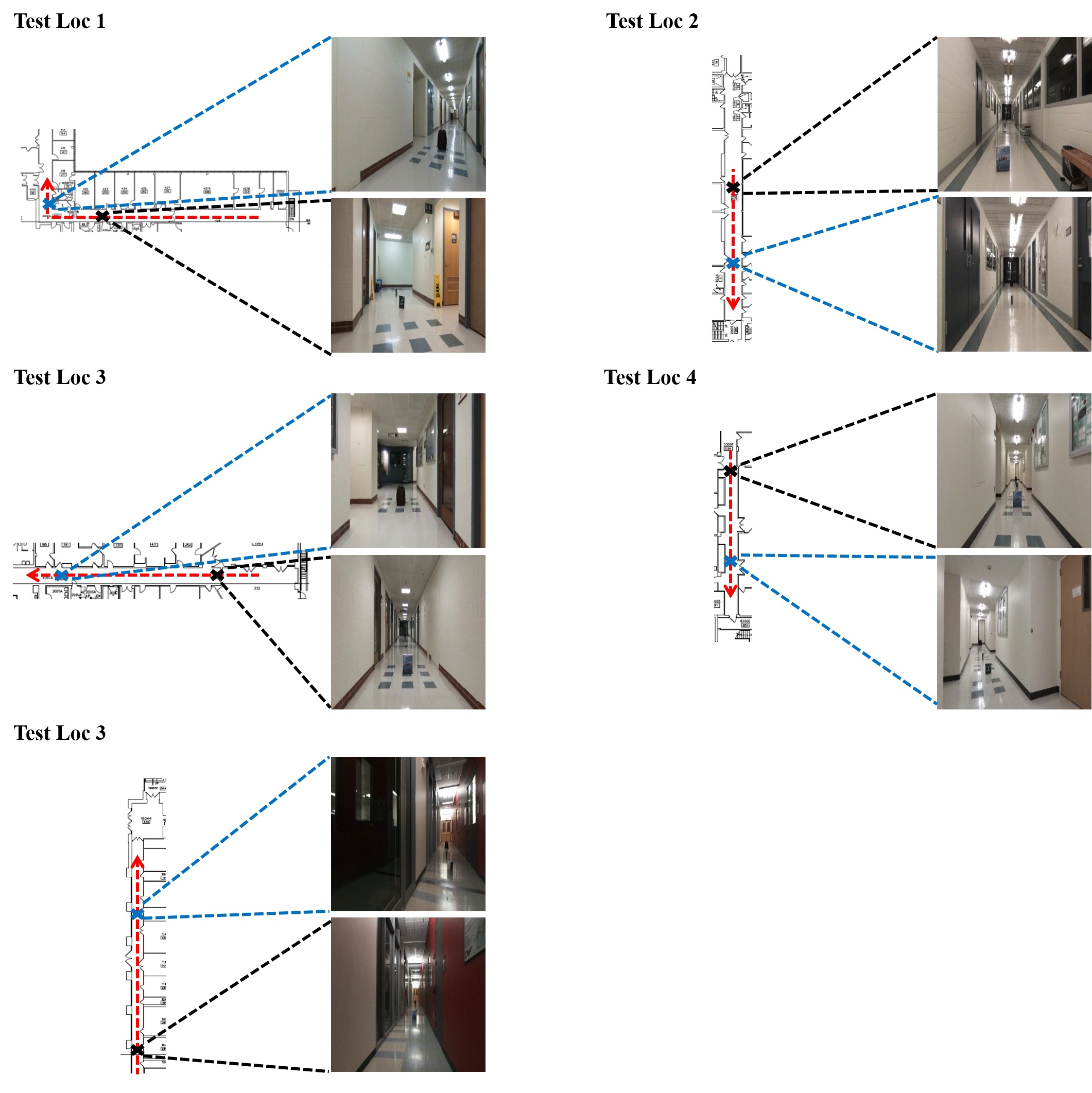}
  \centering
  \caption{Locations for real-time test experiments. For each location, two sample images and a corresponding floor-plan are described. The end of arrow refers to a location where the target is.}
  \label{fig:testFloorPlan}
\end{figure}

\begin{figure}[ht!]
  \includegraphics[scale=0.06]{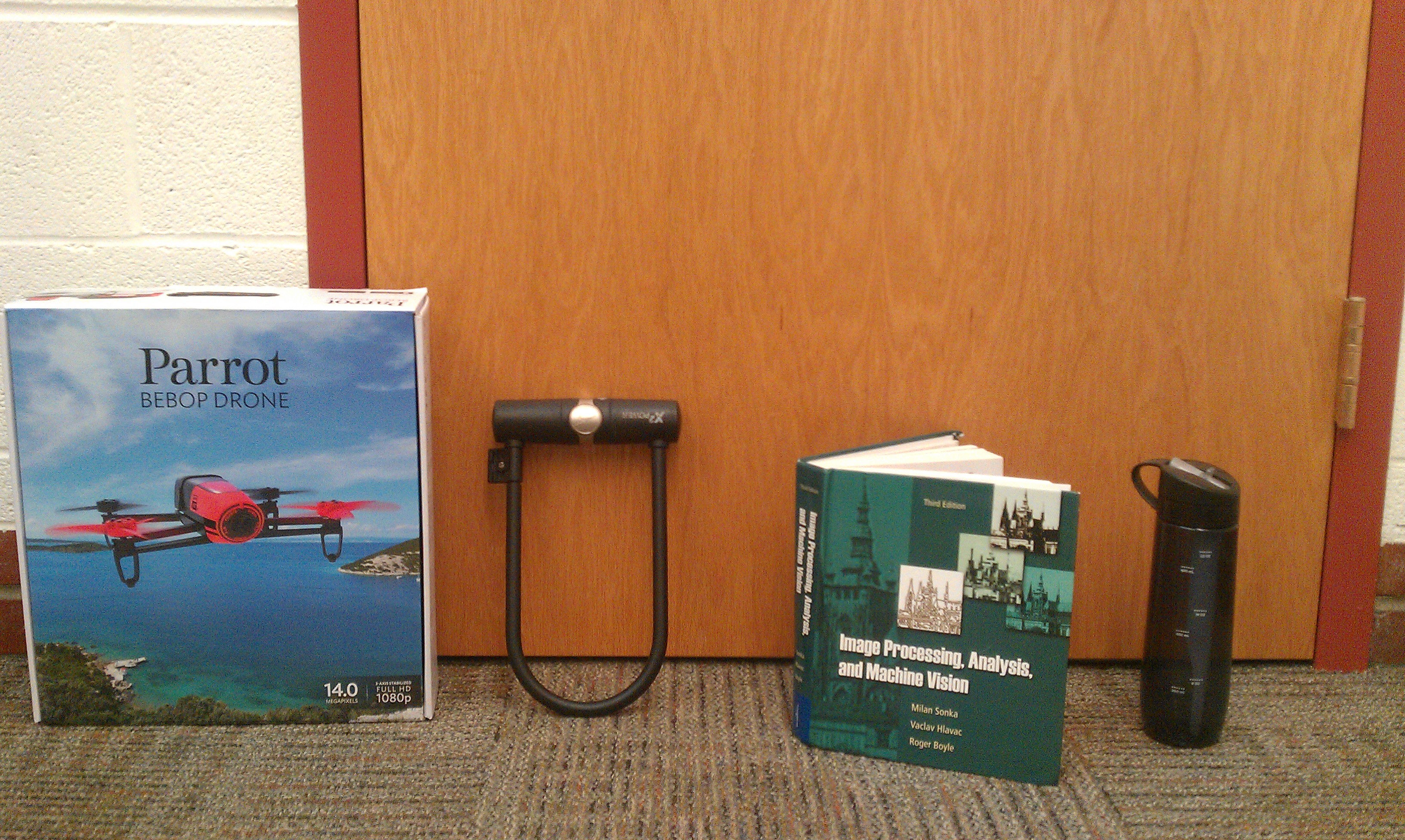}
  \centering
  \caption{Fake targets. From left to right: a box, bike U-lock, book, and water bottle.}
  \label{fig:fake}
\end{figure}

We explain our performance on five test locations. The test locations are chosen to evaluate how well our classifier performs in terms of different objects, geometry, and lighting. The test locations are shown in Figure \ref{fig:testFloorPlan}. Among the five test environments, images from two environments (Test Loc 1 and Test Loc 2) are included in our dataset (Loc 7 and Loc 6), and images from the other three (Test Loc 3, 4, and 5) are not included in our dataset. Test Loc 3 has a similar appearance to the Test Loc 1, but the environment contains different objects (i.e., the glass display stand). Test Loc 4 has different building geometry: it has narrow hallway, compared to other test locations. Test Loc 5 has dim lighting. Test Loc 5 also has the most unique appearance compared to others as its one side is largely composed of glass. To ensure whether our classifier correctly finds the target, we put additional fake targets, a box, book, bike U-lock, and water bottle (Figure \ref{fig:fake}), in the testing environments.\par

Our test policy is described as follows: we count a trial as success only if a quadcopter takes off and flies until it finds a correct target without colliding into any obstacles, i.e. walls. In other words, if quadcopter lands at a wrong target or collide into any obstacles at least once, then we count the trial as a failure. Our test performance is summarized in Table \ref{table:experiment}. The result shows a success rate of 70-80\% for Test Loc 1-4. The classifier has never experienced images taken from Test Loc 3 and 4 but achieved similar performance to the performance in the seen environments. This result suggests that our classifier has a robust performance to fly autonomously in buildings with different objects and geometry. Because Test Loc 5 has the most unique appearance compared to the images in dataset, it shows less success rate of 60\%, but the performance is comparable with the others. The test video could be found at: https://www.youtube.com/watch?v=2Y08GRYnC3U.

\vspace{5cm}
\begin{table}[h!]
\centering
\begin{tabular}{ |c||c| }
 \hline
 Test Environment & Success/Number of Trials\\
 \hline
 Test Loc 1 & 8/10 \\
 Test Loc 2 & 7/10 \\
 Test Loc 3 & 8/10 \\
 Test Loc 4 & 8/10 \\
 Test Loc 5 & 3/5 \\
 \hline
\end{tabular}
\caption[Table caption text]{Real-time test experiment result.}
\label{table:experiment}
\end{table}

\section{Discussion}
Visualization techniques (\cite{yosinski},\cite{deconvnet},\cite{zeiler},\cite{google}) have been proposed to understand deep learning models better. They provides us qualitative inner representations learned by deep networks, and they also allows us to diagnose potential problems with deep models. In this section, we use several visualization techniques (\cite{yosinski},\cite{deconvnet}) to understand more about our trained network.

\subsection{Class Model Visualization}
The objective of class model visualization \cite{yosinski} is to generate a synthetic input image, which causes a high score for a specific class. The resulting synthetic image represents what a trained model is looking for a specific class. We generate the synthetic image by computing a gradient using back-propagation and performing regularizations, such as L2 decay and Gaussian Blur, as described in \cite{yosinski}. We initialize the optimization with the random image. Then we use regularization of L2 decay and apply Gaussian Blur for every four optimization steps.\par
The resulting synthetic images for each class (flight command) is shown in Figure \ref{fig:classVis}. The visualization result suggests that our classifier has learned correct features for each flight command. The Stop flight command visualization, for instance, looks for features of the book bag. The distinction between different class visualization verifies that our network has learned unique features between different flight commands. However, less clear visualizations for Spin Right and Spin Left class suggest that our network could learn better about these classes. Increasing number of images for Spin Right and Spin Left flight command could be one possible solution.

\subsection{Image-Specific Class Saliency Visualization}
\begin{figure}[H]
  \includegraphics[width=\textwidth]{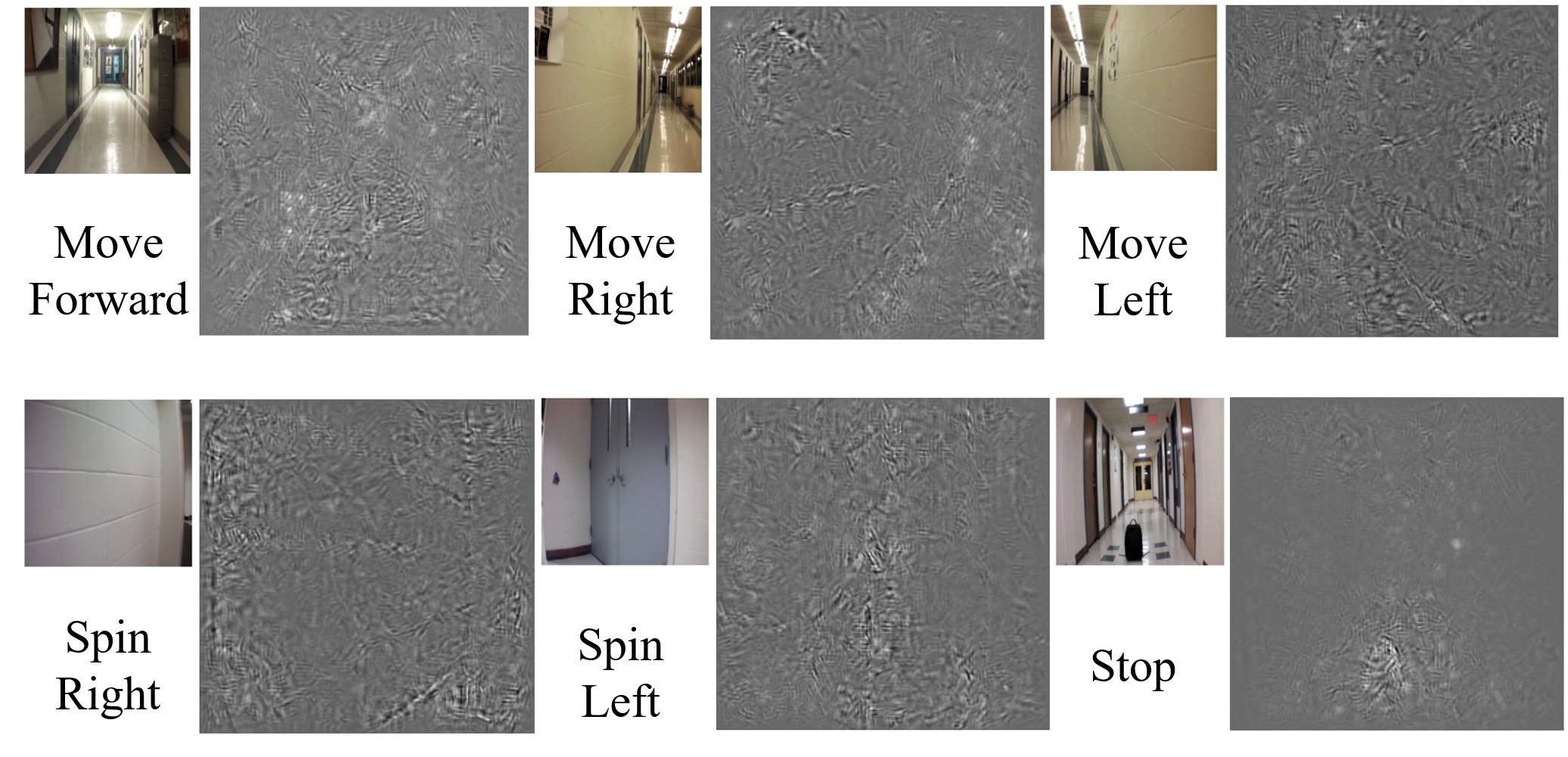}
  \centering
  \caption{Image-specific saliency visualization for each class. The figure indicates which part of an input image affected a class score the most. Brighter (higher intensity) part means that it has larger affect to the class score. The figure is best viewed in color and zoom-in electronic form.}
  \label{fig:imageVis}
\end{figure}

Given an image, an image-specific class saliency map is generated by using the class score derivative and rearranging the derivate \cite{deconvnet}. The saliency map indicates which part of an image affected a class score the most. We select 50 images for each flight command that score the most and generate the saliency maps. Some results for each class is shown in Figure \ref{fig:imageVis}. The results highlight edges in common, which suggests edges are an important feature that affect classification performance.

\afterpage{
\thispagestyle{empty}
\begin{figure}[H]
  \includegraphics[width=\textwidth,height=0.9\textheight,keepaspectratio]{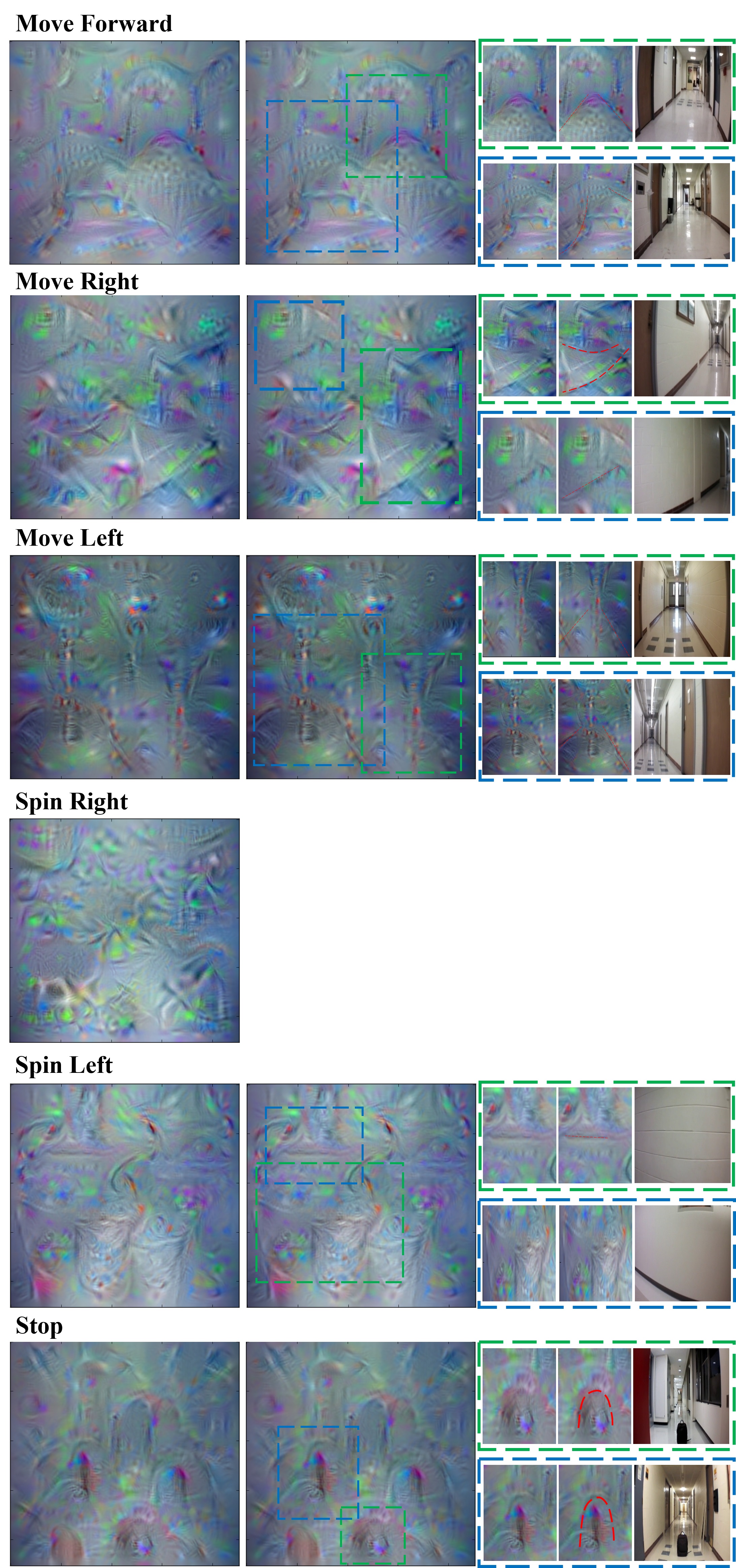}
  \centering
  \caption{Class model visualization of each flight command. The results of visualization are shown in the first column. The visualization result in \cite{yosinski} contains multiple objects that a trained model learned for a certain class. Similarly, our visualization results show multiple indoor environments. For example, in the Move Forward visualization, two indoor environments can be observed, which visualize the same Move Forward flight command. In the second column, we overlay two boxes, a green box and a blue box, to highlight two well visualized indoor environments. We crop each box and show each individually (the first column of each box). We overlay red dotted lines to aid observation (the second column of each box). Then finally, we show an image in the training dataset that is visually most close to the visualization (the third column of each box). Please note that we could not find any well visualized indoor environment in Spin Right visualization. The figure is best viewed in color and zoom-in electronic form.}
  \label{fig:classVis}
\end{figure}
\clearpage
}

\section{Conclusion}
We have presented a deep learning based system that enables a quadcopter to navigate autonomously indoors and find a specific target. Through our real-time experiments, we show that our approach performs well in diverse indoor locations. Our approach is practical as it requires only a single camera and computationally efficient as it does not reconstruct a 3-D map. In this paper, we have investigated our trained network deeper by using visualization techniques. In future work, we want to expand our dataset and experiments to more diverse indoor environments, such as stairways.\par

\section{Acknowledgment}
This work was supported by Engineering Learning Initiatives at Cornell University. We thank Robert A. Cowie for research grant. We thank Hang Chu, Yuka Kihara, Amandianeze Nwana, and Kuan-chuan Peng at Advanced Multimedia Processing laboratory for useful discussions and help. 

\newpage


\end{document}